\documentclass[10pt,twocolumn,letterpaper]{article}

\usepackage{iccv}
\usepackage{times}
\usepackage{epsfig}
\usepackage{graphicx}
\usepackage{subfig}
\usepackage{enumitem}
\usepackage{bm}
\usepackage{mathtools}
\usepackage{siunitx}

\usepackage{algorithm,algcompatible,amssymb,amsmath}

\algnewcommand\algorithmicto{\textbf{to}}
\algnewcommand\RETURN{\State \textbf{return} }

\algnewcommand\algorithmicinput{\textbf{Input:}}
\algnewcommand\INPUT{\item[\algorithmicinput]}

\algnewcommand\algorithmicoutput{\textbf{Output:}}
\algnewcommand\OUTPUT{\item[\algorithmicoutput]}

\algnewcommand\algorithmicinitialize{\textbf{Initialize:}}
\algnewcommand\INITIALIZE{\item[\algorithmicinitialize]}

\usepackage[pagebackref=true,breaklinks=true,letterpaper=true,colorlinks,bookmarks=false]{hyperref}

\iccvfinalcopy 

\def\iccvPaperID{3596} 

\ificcvfinal\fi

\begin{document}

\title{Learning Spatial Awareness to Improve Crowd Counting}

\author{Zhi-Qi Cheng$^{1,2}$\thanks{\footnotesize indicates equal contribution.  This work was done when Zhi-Qi Cheng and Jun-Xiu Li were visiting at Microsoft Research. Xiao Wu is the corresponding author.}, ~ Jun-Xiu Li$^{1,3}$\footnotemark[1], ~ Qi Dai$^3$,  ~ Xiao Wu$^1$,  ~Alexander G. Hauptmann$^2$ \\
	{$^1$Southwest Jiaotong University, 
		$^2$Carnegie Mellon University, 
		$^3$Microsoft Research}\\
	\tt\footnotesize \{zhiqic, alex\}@cs.cmu.edu,
	\tt\footnotesize  \{lijunxiu@my, wuxiaohk@home\}.swjtu.edu.cn,
	\tt\footnotesize qid@microsoft.com\\
}
\maketitle
\ificcvfinal\thispagestyle{empty}\fi

\begin{abstract}
\vspace{-4mm}
 The aim of crowd counting is to estimate the number of people in images by leveraging the annotation of center positions for pedestrians' heads. 
 Promising progresses have been made with the prevalence of deep Convolutional Neural Networks.
 Existing methods widely employ the Euclidean distance (i.e., $L_2$ loss) to optimize the model, which, however, has two main drawbacks: (1) the loss has difficulty in learning the spatial awareness (i.e., the position of head) since it struggles to retain the high-frequency variation in the density map, and (2) the loss is highly sensitive to various noises in crowd counting, such as the zero-mean noise, head size changes, and occlusions.
 Although the Maximum Excess over SubArrays (MESA) loss has been previously proposed by~\cite{nips-10} to address the above issues by finding the rectangular subregion whose predicted density map has the maximum difference from the ground truth, it cannot be solved by gradient descent, thus can hardly be integrated into the deep learning framework.
 In this paper, we present a novel architecture called SPatial Awareness Network (SPANet) to incorporate spatial context for crowd counting.
 The Maximum Excess over Pixels (MEP) loss is proposed to achieve this by finding the pixel-level subregion with high discrepancy to the ground truth.
 To this end, we devise a weakly supervised learning scheme to generate such region with a multi-branch architecture.
 The proposed framework can be integrated into existing deep crowd counting methods and is end-to-end trainable.
 Extensive experiments on four challenging benchmarks show that our method can significantly improve the performance of baselines.
 More remarkably, our approach outperforms the state-of-the-art methods on all benchmark datasets.
\end{abstract}

\vspace{-4mm}
\section{Introduction}
\label{sec:introduction}
\vspace{-1mm}
The problem of crowd counting is described in~\cite{nips-10}.
Different from visual object detection, it is impossible to provide bounding boxes for all pedestrians due to the extremely dense crowds.
On the other side, when only the total crowd counts of the images are provided, the training process will become notably difficult since the spatial awareness is completely ignored. Therefore, to preserve as many spatial constraints as possible and reduce annotation cost, the previous work~\cite{nips-10} started to only provide center points of heads and utilizes Gaussian distribution to generate ground truth density maps. It is worth noting that this annotation scheme is widely adopted by subsequent studies.

Existing crowd counting approaches mainly focus on improving the scale invariance of feature representation, including the multi-column networks~\cite{deep-multi-attention-AFP-18,deep-fscale-multi-task-17,deep-multi-pyramid-cnn-17,deep-multi-cnn-boosting-16, MCNN-16,MM-2018}, scale aggregation modules~\cite{SANet-18,deep-multi-block-mscnn-17}, and scale-invariant networks~\cite{SPooling-18,CSRNet-18,deep-attention-cnn-ADCrowdNet-18,deep-multi-pyramid-cnn-17,deep-pooling-fscale-Defense-18}. Despite the architectures of these methods are different, the $L_2$ loss function is employed by most of them. 
As a result, the spatial awareness in crowd image is largely ignored, though more scale information is embedded into their features.

\begin{figure}[t]
	\footnotesize
	\begin{center}
		\includegraphics[width=0.95\linewidth]{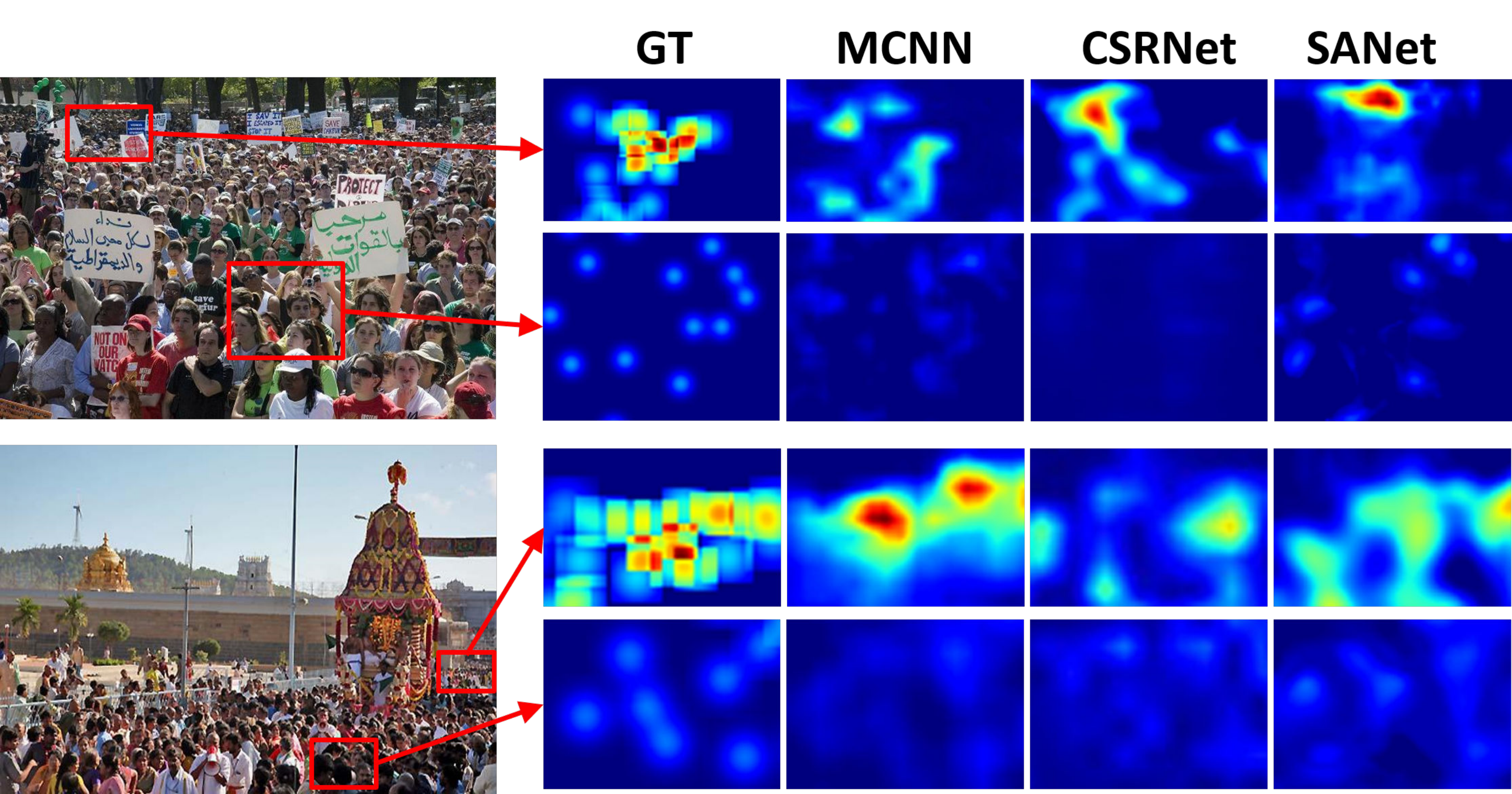}
	\end{center}
	\vspace{-5mm}
	\caption{\footnotesize The $L_2$ loss function has difficulty in learning the spatial awareness and is sensitive to various noises in crowd counting,
	which will lead to a lower estimation in high-density regions (the first row of each example), and a higher estimation in low-density regions (the second row of each example). 
	Note that the corresponding improvements of our method are shown in Figure \ref{fig:qualitative}.
}
	\label{fig:l2-problem-1}
	\vspace{-6mm}
\end{figure}

We have examined three state-of-the-art approaches (i.e.,~MCNN~\cite{MCNN-16}, CSRNet~\cite{CSRNet-18}, and SANet~\cite{SANet-18}) on four crowd counting datasets (i.e.,~ShanghaiTech~\cite{MCNN-16}, UCF\_CC\_50~\cite{r-mrf-count-multi-source-13}, WorldExpo'10~\cite{deep-crowd-scene-15}, and UCSD~\cite{r-count-privacy-preserving-08}). 
Two examples are shown in Figure \ref{fig:l2-problem-1}.
Similar to \cite{SANet-18,deep-decide-18,deep-attention-cnn-ADCrowdNet-18}, we observe that dense-crowd regions are usually underestimated, while sparse-crowd regions are overestimated. Such phenomenon is due to two main factors.
First, the pixel-wise $L_2$ loss struggles to retain the high-frequency variation in the density map: minimizing $L_2$ loss encourages finding pixel-wise averages of plausible solutions which are typically overly-smooth and thus have poor spatial awareness \cite{ledig2017photo}.
Second, $L_2$ loss is highly sensitive to typical noises in crowd counting, including the zero-mean noise, head size changes, and head occlusions. We take a simple statistics and show that the co-occurrence of zero-mean noise and overestimation could reach 96\% (6,776 out of 7,044 testing images). We further find that almost all estimated density maps inaccurately predict the head positions or sizes when occlusion occurs, which could result in underestimation in high-density areas. Moreover, the generated ground truth density could also be imprecise due to the annotation error and the fixed variance in Gaussian kernel.
It is noted that the corresponding improvements of our method are illustrated in Figure \ref{fig:qualitative}.

To fully utilize the spatial awareness, previous work~\cite{nips-10} proposes a loss named Maximum Excess over SubArrays (MESA) to handle the above problems.
Generally speaking, MESA loss attempts to find the rectangular subregion whose predicted density map has the maximum difference from the ground truth. It directly optimizes the counts of this subregion instead of the pixel-level density. Since the set of subregions could include the full image, MESA loss is an upper bound for the count estimation of the entire image.
Besides, this loss is only sensitive to the spatial layout of pedestrians and is robust to various noises.
However, the complexity of MESA loss function is extremely high.
\cite{nips-10} utilizes Cutting-Plane optimization to obtain an approximate solution. Since this method cannot be solved by the conventional gradient descent, MESA loss has not been employed in any existing CNN-based approach.

Motivated by the MESA loss, in this paper we present a novel deep architecture called SPatial Awareness Network (SPANet) to retain the high-frequency spatial variations of density.
Instead of finding the mismatched rectangular subregion as in MESA, the Maximum Excess over Pixels (MEP) loss is proposed to optimize the pixel-level subregion which has high discrepancy to the ground truth density map.
To obtain such pixel-level subregion, the weakly-supervised ranking information \cite{L2Rank-18} is exploited to generate a mask indicating the pixels with high discrepancies.
We further devise a multi-branch architecture to leverage the full image for discrepancy detection by imitating the salience region detection \cite{grad-17,agad-18,cam-16}, where patches with increasing areas are used for ranking.
The proposed framework could be easily integrated into existing CNN-based methods and is end-to-end trainable.

The main contribution of this work is the proposed Spatial Awareness Network and Maximum Excess over Pixels loss for addressing the issue of crowd counting.
The solution also provides the elegant views of what kind of spatial context should be exploited and how to effectively utilize such spatial awareness in crowd images, which are problems not yet fully understood in the literature.

\section{Related Work}
\vspace{-1mm}
\subsection{Detection-based Methods}
\vspace{-1mm}
The methods in this category use object detector to locate people in images. Given the individual localization of each people, crowd counting becomes trivial. 
There are two directions in this line, i.e.,~detection on 1) whole pedestrians \cite{d-bayesian-full-06,d-histograms-svm-full-05,d-full-bayesian-08} and 2) parts of pedestrians \cite{d-part2-08,d-part-15,d-harr-svm-part-01,d-part-11}. 
Typically, local features \cite{d-histograms-svm-full-05,d-harr-svm-part-01} are first extracted and then are exploited to train various detectors (e.g., SVM~\cite{d-harr-svm-part-01} and AdaBoost~\cite{d-adaboost-full-05}).
Though spatial information is well learned in these methods, they are not applicable in challenging situations, such as the high-density clogging crowds.

\vspace{-1mm}
\subsection{Regression-based Methods}
\vspace{-1mm}
Different from detection-based methods, regression-based approaches avoid the hard detection problem and estimate crowd counts from image features.
Earlier methods \cite{r-count-privacy-preserving-08,r-edge-count-bayesian-12,r-mrf-count-multi-source-13,r-edge-count-96} usually predict the counts directly from the features, which will lead to poor performance as the spatial awareness is completely ignored. 
Later methods try to estimate the density map for counting \cite{nips-10,r-forest-density-15,r-density-edge-09}, where the crowd count is obtained by integrating all pixel values over the density map.
Though learning the density map somewhat provides the spatial information, their models still have difficulties in preserving the high-frequency variation in the density map.

\vspace{-1mm}
\subsection{CNN-based Methods}
\vspace{-1mm}
Deep CNN based crowd counting methods have shown very strong performance improvements over the shallow learning counterparts. Existing methods mainly focus on coping with the large variation in pedestrian scales, where many multi-column networks are extensively studied.
A dual-column network is proposed by~\cite{deep-shallow-deep-crowdnet-16} to combine shallow and deep layers for estimating the count. Inspired by this work, a famous three-column network MCNN is proposed by~\cite{MCNN-16}, which employs different filters on separate columns to obtain features with various scales. Many works have improved MCNN~\cite{deep-multi-attention-AFP-18,deep-fscale-multi-task-17,deep-multi-pyramid-cnn-17,deep-multi-cnn-boosting-16} to further enhance the scale adaptation. Sam \emph{et al.}~\cite{deep-switch-17} introduce a switching structure, which uses a classifier to assign input image patches to appropriate columns. Recently, Liu \emph{et al.}~\cite{deep-decide-18} propose a multi-column network to simultaneously estimate crowd density by detection and regression based models. Ranjan \emph{et al.}~\cite{deep-training-ICCNN-18} utilize a two-column network to iteratively train their model with images of different resolution. 

There are a lot of other attempts to further improve the scale invariance, including 1) study on the fusion of various scale information~\cite{deep-fscale-Context-18,deep-fscal-pooling-PaDNet-18,deep-pooling-fscale-Defense-18,deep-multi-fusion-Adaptive-18}, 
2) study on multi-blob based scale aggregation networks~\cite{SANet-18,deep-multi-block-mscnn-17}, 
3) design of scale-invariant convolutional or pooling layers~\cite{SPooling-18,CSRNet-18,deep-attention-cnn-ADCrowdNet-18,deep-multi-pyramid-cnn-17,deep-pooling-fscale-Defense-18}, 
and 4) study on the automated scale adaptive networks~\cite{deep-auto-TDF-CNN-18,deep-auto-IGCNN-18,deep-fscale-multitask-SaCNN-18}.
Typically, Li \emph{et al.}~\cite{CSRNet-18} propose CSRNet that exploits dilated convolutional layers to enlarge receptive fields for boosting performance. Cao \emph{et al.}~\cite{SANet-18} propose SANet to aggregate multi-scale features for more accurate crowd count. These two approaches have achieved state-of-the-art performance. Additionally, there also exist studies devoted to utilization of perspective maps~\cite{Perspective-18}, geometric constraints~\cite{deep-Geometric-18,deep-geo-Att-head-18}, and region-of-interest (ROI)~\cite{deep-attention-cnn-ADCrowdNet-18} to improve the counting accuracy.

The aforementioned methods utilize the Euclidean distance, i.e.~$L_2$ loss to optimize the model. Although these methods can obtain scale-invariant features, their performances are still unsatisfactory since the spatial awareness is largely ignored.
Note that, SANet~\cite{SANet-18} also tries to solve the problem of $L_2$ loss and adds local pattern consistency ($L_c$ loss) in the training phase.
However, we find that $L_c$ still cannot learn the spatial context well. 
In our experiment, when integrating our MEP loss ($L_{mep}$) into SANet, we achieve significant performance improvement. 
Our proposed MEP loss could fully utilize the spatial awareness, which is a key factor for the task of crowd counting.

\vspace{-1mm}
\section{Our Method}
\vspace{-1mm}
In this section, we first review the problem of crowd counting and
two loss functions (i.e.,~MESA loss and $L_2$ loss).
Then we present the proposed SPANet and MEP loss in details. 
It is worth noting that our method can be directly applied to
all CNN-based crowd counting networks.
\vspace{-1mm}
\subsection{Problem Formulation}
\vspace{-1mm}
\label{sec:revisiting-of-crowd-counting}
Recent technologies define the crowd counting task as a density regression problem~\cite{SANet-18,nips-10,MCNN-16}.
Given $N$ images $\textbf{I}=\{I_{1}, I_{2}, \cdots, I_{N}\}$ as the training set, each image $I_{i}$ is annotated with a total of $c_{i}$ center points of pedestrians' heads $\textbf{P}_{i}^{gt}=\{P_{1}, P_{2}, \cdots, P_{c_{i}}\}$.
Typically, the ground truth density map for each pixel $p$ in image $I_{i}$ is defined as $D^{gt,i}$,
\begin{equation}
\footnotesize
\label{eqn:ground-truth-density}
\forall p\in I_{i}, D^{gt,i}(p)=\sum_{P\in\textbf{P}_{i}^{gt}}\mathcal{N}^{gt}(p;\mu=P,\sigma^{2}),
\vspace{-2mm}
\end{equation}
where $\mathcal{N}^{gt}$ is a Gaussian distribution. The number of people $c_{i}$ in image $I_{i}$ is equal to the sum of density values over all pixels as $\sum_{p\in I_{i}}D^{gt,i}(p)=c_{i}$. With these training data, the aim of crowd counting task is to learn the predicted density map $D^{pr}$ towards the ground truth density map $D^{gt}$.
\begin{figure}[!thb]
	\footnotesize
	\begin{center}
		\includegraphics[width=0.6\linewidth]{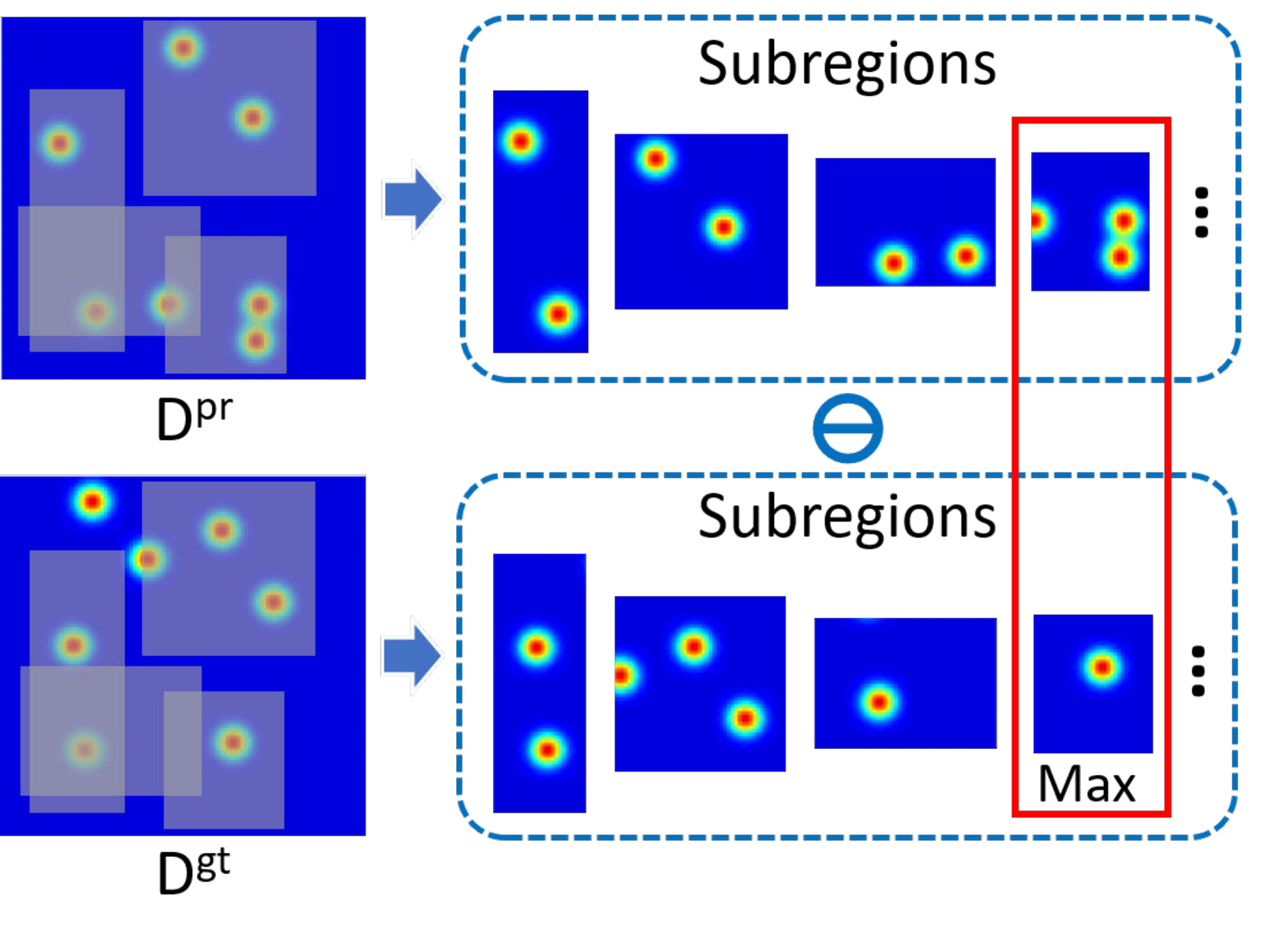}
	\end{center}
	\vspace{-8mm}
	\caption{\footnotesize Computation process of MESA loss. It is required to traverse all possible subregions and calculate the differences between their predicted density maps and the ground truth. Then the subregion with maximum difference is selected for optimization.}
	\label{fig:l-mesa}
	\vspace{-6mm}
\end{figure}

\textbf{MESA loss}.
To make use of the spatial awareness in annotations (i.e.,~center head positions $\textbf{P}^{gt}$), the previous work~\cite{nips-10} has proposed the Maximum Excess over SubArrays (MESA) loss $L_{mesa}$ as follows,
\begin{equation}
\footnotesize
\label{eqn:l-mesa}
L_{mesa}\left(D^{pr},D^{gt}\right)=\frac{1}{N}\sum_{i=1}^{N}\max_{B \in \mathbf{B}} \left | \sum_{p \in B} D^{pr,i}\left(p\right) - \sum_{p \in B} D^{gt,i}\left(p\right) \right |,
\end{equation}
where $\mathbf{B}$ is the set of all potential rectangular subregions in image. 
As illustrated in Figure~\ref{fig:l-mesa}, MESA loss tries to find the box subregion whose predicted density map has the maximum difference from the ground truth. It can be treated as an upper bound for the count estimation of the entire image, as $\mathbf{B}$ could include the full image. Besides, this loss is directly related to the counting objective instead of the pixel-level density, and is only sensitive to the spatial layout of pedestrians. In the 1D case, Kolmogorov-Smirnov distance~\cite{kolmogorov-51} can be seen as a special case of $L_{mesa}$.

Despite the above merits, it is difficult to optimize the MESA loss due to the hard process of finding such subregion. One has to traverse all potential subregions to achieve this, which is obviously an impossible task in practical application.
To solve it, previous approach \cite{nips-10} converts the optimization of MESA loss to a convex quadratic program problem with limited constraints and utilizes Cutting-Plane optimization to obtain an approximate solution. However, since this method cannot be solved by the traditional gradient descent, MESA loss has not been exploited in any existing CNN-based crowd counting methods.

\begin{figure*}[!thb]
	\footnotesize
	\begin{center}
		\includegraphics[width=0.94\textwidth]{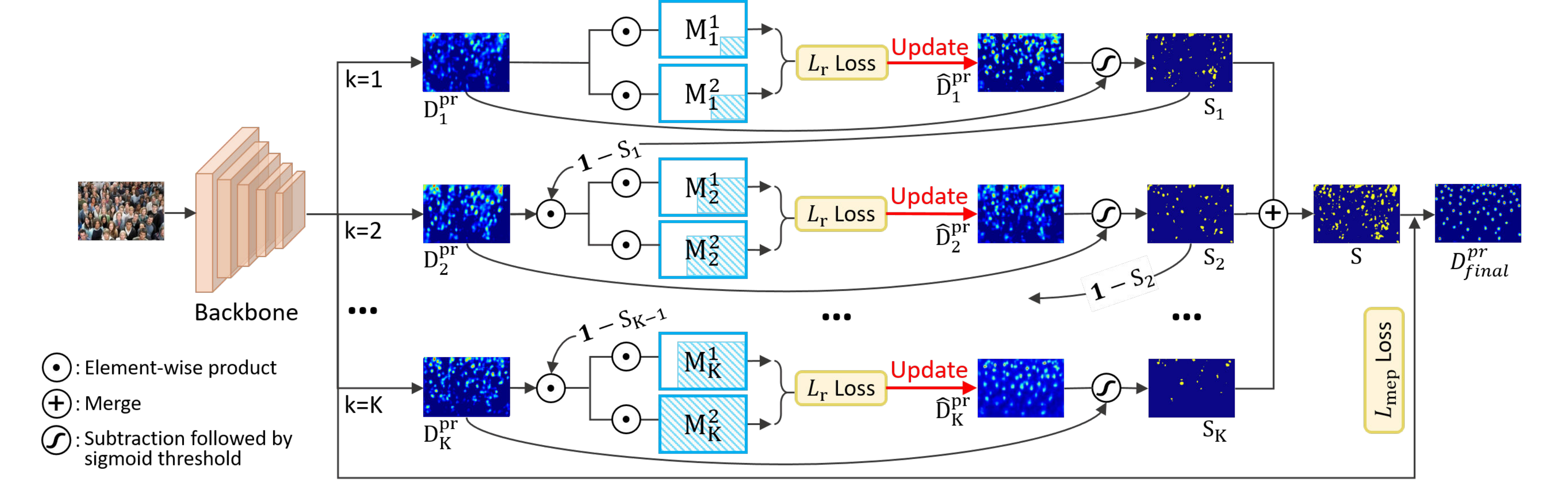}
	\end{center}
	\vspace{-4mm}
	\caption{\footnotesize The framework of our proposed SPatial Awareness Network (SPANet). The input images are first fed into the backbone network to extract feature representations and output the estimated density maps $D^{pr}$. A $K$-branch architecture is devised. In each branch $k$, the network is optimized with the ranking objective by sampling two patches (one is sub-patch of the other) and outputs a new density map $\hat{D}^{pr}_k$. Then the two density maps are utilized to produce the subregion $S_k$ which has high discrepancy to the ground truth.
		The density values within the generated $S_k$ is erased in next branch to facilitate the latter optimization.
		In the end, $K$ subregions from $K$ branches are fused to form the final pixel-level subregion $S$, which is exploited to calculate the Maximum Excess over Pixels (MEP) loss.}
	\label{fig:framework}
	\vspace{-4mm}
\end{figure*}

\textbf{$\bm{L_{2}}$ loss}.
To facilitate the computation in deep frameworks, existing CNN-based methods~\cite{CSRNet-18,deep-training-ICCNN-18,MCNN-16} all directly use $L_2$ loss to minimize the difference between the estimated and ground truth density maps,
\begin{equation}
\footnotesize
L_2\left(D^{pr},D^{gt}\right)=\frac{1}{2N}\sum_{i=1}^{N}\sum_{p \in D^{pr,i}} \left | \left |  D^{pr,i}\left(p\right) -  D^{gt,i}\left(p\right) \right | \right |^{2}_{2}.
\end{equation}
However, as discussed in Sec.~\ref{sec:introduction}, we reveal that $L_2$ loss can hardly retain the high-frequency variation in the density map, leading to the poor spatial awareness. Furthermore, it is also highly sensitive to typical noises in crowd counting, including the zero-mean noise, head size changes, and head occlusions.
For example, existing methods always overestimate the density value in low-density areas and underestimate it in high-density regions.

\subsection{Spatial Awareness Network}
\vspace{-1mm}
\label{sec:optimized-l-mesa}
The proposed Spatial Awareness Network (SPANet) aims to leverage the spatial context for accurately predicting the density values.
Instead of searching the mismatched rectangular subregion as in MESA loss, which is the main obstacle for optimization, we try to find the pixel-level subregion ${S}$ which has high discrepancy to the ground truth density map.
Since there is not any annotation of such region, this problem is unsupervised and will still be significantly difficult to solve.
Inspired by the recent weakly-supervised method \cite{L2Rank-18}, we exploit an obvious ranking relation to achieve this, i.e., one patch of a crowded scene image is guaranteed to contain the same number or fewer persons than the original image.
By sampling a pair of patches (where one is the sub-patch of the other), the network is optimized with the ranking objective and outputs a new density map, which is in turn utilized to produce the subregion with high discrepancy, together with the previous one.
We further devise a multi-branch architecture to leverage the full image by sampling multiple pairs of patches.
Note that the whole SPANet could be end-to-end trained.

Figure~\ref{fig:framework} illustrates the framework of our proposed SPANet.
Input images $I$ are first fed into the backbone network 
to generate the predicted density maps $D^{pr}$.
The desired \textbf{\emph{pixel-level subregion generation}}, i.e., $S_k$, is conducted by branch $k$ using a pair of patches sampled from density maps $D^{pr}$.
To leverage the full image for discrepancy detection, a \textbf{\emph{multi-branch architecture}} with $K$ branches is devised to produce multiple subregions by imitating the salience region detection \cite{agad-18,cam-16}.
Finally, $K$ subregions ($S_1, S_2, ..., S_K$) are combined to produce the final ${S}$, which is then exploited to compute our proposed \textbf{\emph{Maximum Excess over Pixels (MEP) loss}}.
We will present these three sub-modules in details below.

\textbf{Pixel-level Subregion Generation}.
The subregion ${S}$ indicates the area with high density discrepancy to the ground truth.
Unfortunately, directly subtracting the predicted $D^{pr}$ from the ground truth $D^{gt}$ would make the problem go round in circles -- the bias is usually large enough to prevent it from providing accurate region.
Consequently, we turn to find the region with high changes along with the network training.
It is natural that one can pick two density maps of the same image from different iterations.
However, the obtained area only reflects the region that is already ``revised'', which still seriously suffers from the poor spatial perception of the original $L_2$ loss.
To this end, we exploit the weakly supervised ranking clues to produce the subregion.
Instead of considering the pixel-level density, the ranking clue is directly related to the comparison of crowd counts.


In each branch $k$, two parallel image patches are first sampled.
As the feature maps of deep convolutional layers already contain rich location information, we treat the sampling process as the mask pooling operation on the density map.
The strategy of selecting patches will be described later.
Without loss of generality, suppose the two masks $M_k^1$ and $M_k^2$ are the 2-dimensional matrix with $0$ or $1$ ($1$ indicates the patch area), and $M_k^1$ is the sub-patch of $M_k^2$.
The crowd counts $C(M_{k}^1)$ and $C(M_{k}^2)$ under the masks $M_k^1$ and $M_k^2$ could be obtained by integrating the values of density map over individual mask, which could be implemented as the mask pooling as follows,
\begin{equation}
\footnotesize
\label{eqn:crowd-count}
\vspace{-1mm}
\begin{aligned}
C\left(M_{k}^1\right)=\sum_{p \in D^{pr}_k}\left(D^{pr}_k \odot  M_k^1\right),\\
C\left(M_{k}^2\right)=\sum_{p \in D^{pr}_k}\left(D^{pr}_k \odot  M_k^2\right),
\vspace{-1mm}
\end{aligned}
\end{equation}
where $\odot$ is the element-wise product, and $p$ indicates the pixel on density map $D^{pr}_k$. 
It is worth noting that we utilize the same predicted density map $D^{pr}_k$ when calculating the counts for two masks, rather than generating individual maps at two consecutive iterations.
The reason is that the density map $D^{pr}_k$ is not restricted to be positive, thus pooling on the pair of patches could also provide the ranking information.
We have conducted an experiment showing that the two schemes have similar results.
Besides, directly pooling on the same map is more efficient than the other.

With the assumption that $M_k^1$ is the sub-patch of $M_k^2$, the explicit constraint is that the number of people in $M_k^1$ is fewer than that in $M_k^2$.
Therefore, we employ a pairwise ranking hinge loss $L_r$ to model such relationship, which is formulated as
\begin{equation}
\small
\label{eqn:rank}
L_{r}\left({C}(M_{k}^1),{C}(M_{k}^2)\right)=\max\left(0,~{C}(M_{k}^1)-{C}(M_{k}^2)+\xi \right),
\end{equation}
where $\xi$ is a margin value that is set to the upper bound of the difference in the ground truth.
The gradient of $L_{r}$ loss is calculated as
\begin{equation}
\footnotesize
\label{eqn:gradient}
\triangledown_\theta  L_{r}=\left\{\begin{matrix}
0, \;\;\;\;\;\;\;\;\;\;\; \mbox{if\;\;} {C}\left(M_{k}^1\right)- {C}\left(M_{k}^2\right)+\xi \leqslant 0, \\ 
\\
\triangledown_\theta {C}\left(M_{k}^1\right)- \triangledown_\theta {C}\left(M_{k}^2\right),  \;\;\;\;\;\;\; \mbox{otherwise}.
\end{matrix}\right.
\vspace{-1mm}
\end{equation}

Once the network parameters $\theta$ are updated with $L_r$ by back-propagation, the renewed density map $\hat{D}^{pr}_k$ estimated by the network is computed by
\begin{equation}
\small
\label{eqn:renewed-density}
\hat{D}^{pr}_k=\mbox{Conv}\left(I,\theta \right),
\end{equation}
where $I$ is the input image, and $\mbox{Conv}(\cdot)$ refers to a forward pass of the network. 
Given the updated density map $\hat{D}^{pr}_k$ and the old one ${D}^{pr}_k$, the desired subregion $S_k$ is obtained by thresholding the difference $\triangledown D^{pr}_k$ between them, where $\triangledown D^{pr}_k=|\hat{D}^{pr}_k-{D}^{pr}_k|$.
To make it differentiable, we utilize a Sigmoid thresholding function, and $S_k$ is given by
\begin{equation}
\small
\label{eqn:threshold}
S_k=\frac{1}{1+\mbox{exp}\left(-\delta\left(\triangledown D^{pr}_k-\mathbf{\Sigma}\right)\right )},
\end{equation}
where $\mathbf{\Sigma}$ is a threshold matrix with all elements being $\sigma$. 
$\delta$ is the parameter to ensure that the value of $S_k$ is approximately equal to $1$ when $\triangledown D^{pr}_k(p) > \sigma$, otherwise $0$. 

\textbf{Multi-branch Architecture}.
Note that in above section, only a pair of patches are sampled for generating the subregion.
In principle, we hope that the full density map could be leveraged to provide more information.
Instead of only sampling a small-large pair of patches, which may involve large bias error due to the large difference between two patches, we adopt a multi-branch architecture as shown in Figure \ref{fig:framework}. 
The bottom right corners of all patches are located at the same position, i.e., the bottom right corner of the density map.
The area of patch is gradually enlarged along with the branches, until it reaches the size of full density map.
Such design guarantees both the small bias error in each branch and the full utilization of training images.

To eliminate the influence of the detected subregion $S_k$ for better optimization in latter branches, we imitate the salience region detection \cite{agad-18} to erase the density values within $S_k$ in next branch, which is formulated as
\begin{equation}
\small
\label{eqn:mask-1}
{D}^{pr}_{k+1}={D}^{pr}_{k+1}\odot(\bm{1}-S_k),
\end{equation}
where $\bm{1}$ is the matrix with all elements being $1$, and $\odot$ is the element-wise product.

\textbf{Maximum Excess over Pixels (MEP) loss}.~In the end, $K$ subregions ($S_1, S_2, ..., S_K$) are generated by the $K$ branches. 
The final desired pixel-level subregion $S$ is computed by simply combining them together as
\begin{equation}
\small
\label{eqn:max-sub-region}
{S}=\sum_{k=1}^{K}\left\{S_k\right\},
\end{equation}
where $\sum$ indicates merging pixels with values close to 1 in all subregion masks $\{S_k\}$, rather than the direct summation.
In practice, we take the maximum value at each pixel position from all masks. The final output $S$ is the mask that indicates the pixels which should be optimized.
Based on that, our proposed MEP loss is then given by 
\begin{equation}
\footnotesize
\label{eqn:l-mep}
L_{mep} \left(D^{pr},D^{gt}\right)=\frac{1}{N}\sum_{i=1}^{N} \left |  \sum_{p\in S} D^{pr,i}(p) -  \sum_{p\in S} D^{gt,i}(p)\right |.
\end{equation}

\subsection{Model Learning}
\vspace{-1mm}
\label{sec:model-learning}

Our SPANet could be easily integrated into existing crowd counting methods, which is equivalent to adding a pooling layer with different masks on the final convolutional layer.
It is trained by sequentially optimizing the $K$ times ranking loss, MEP loss, and the original loss of existing methods.
When calculating the original loss, the mask pooling layer is removed.
The overall training objective is formulated as
\begin{equation}
\footnotesize
\label{eqn:global-loss}
L_{global}= \sum_{k=1}^{K}L_r+L_{mep}+L_{vanilla},
\end{equation}
where $L_{vanilla}$ refers to the original loss of existing approach. In most cases, $L_{vanilla}$ is the $L_2$ loss.
More details of the ground truth generation and data augmentation are described in supplementary material.


\begin{table*}
	\footnotesize
	\begin{center}
		\caption{\footnotesize Performance comparison with the state-of-the-art methods on ShanghaiTech~\cite{MCNN-16}, UCF\_CC\_50~\cite{r-mrf-count-multi-source-13}, and UCSD~\cite{deep-crowd-scene-15} datasets.}
		\vspace{-0.1in}
		\label{tab:stoa-1}
		\begin{tabular}{|lll|cc|cc|cc|cc|}
			\hline
			&                       &                        & \multicolumn{2}{c|}{\textbf{ShanghaiTech A}} & \multicolumn{2}{c|}{\textbf{ShanghaiTech B}} & \multicolumn{2}{c|}{\textbf{UCF\_CC\_50}} & \multicolumn{2}{c|}{\textbf{UCSD}}     \\ \hline
			\multicolumn{1}{|l|}{\textbf{Method}} & \multicolumn{2}{l|}{\textbf{Venue \& Year}}     & \textbf{MAE} $\downarrow$               & \textbf{MSE} $\downarrow$              & \textbf{MAE} $\downarrow$              & \textbf{MSE} $\downarrow$               & \textbf{MAE} $\downarrow$              & \textbf{MSE} $\downarrow$             & \textbf{MAE} $\downarrow$            & \textbf{MSE} $\downarrow$            \\ \hline \hline
			\multicolumn{1}{|l|}{Idrees et al.~\cite{r-mrf-count-multi-source-13}}   & CVPR                  & 2013                   & -                & -               & -               & -                & 419.5           & 541.6          & -             & -             \\
			\multicolumn{1}{|l|}{Zhang et al.~\cite{deep-crowd-scene-15}}    & CVPR                  & 2015                   & 181.8            & 277.7           & 32.0            & 49.8             & 467.0           & 498.5          & 1.60          & 3.31          \\
			\multicolumn{1}{|l|}{CCNN~\cite{deep-multi-perspetive-free-16}}            & ECCV                  & 2016                   & -                & -               & -               & -                & -               & -              & 1.51          & -             \\
			\multicolumn{1}{|l|}{Hydra-2s~\cite{deep-multi-perspetive-free-16}}        & ECCV                  & 2016                   & -                & -               & -               & -                & 333.7           & 425.3          & -             & -             \\
			\multicolumn{1}{|l|}{C-MTL~\cite{deep-fscale-multi-task-17}}           & AVSS                  & 2017                   & 101.3            & 152.4           & 20.0            & 31.1             & 322.8           & 397.9          & -             & -             \\
			\multicolumn{1}{|l|}{SwitchCNN~\cite{deep-switch-17}}       & CVPR                  & 2017                   & 90.4             & 135.0           & 21.6            & 33.4             & 318.1           & 439.2          & 1.62          & 2.10          \\
			\multicolumn{1}{|l|}{CP-CNN~\cite{deep-multi-pyramid-cnn-17}}          & ICCV                  & 2017                   & 73.6             & 106.4           & 20.1            & 30.1             & 295.8           & 320.9          & -             & -             \\
			\multicolumn{1}{|l|}{Huang at al.~\cite{deep-body-BSAD-18}}    & TIP                   & 2018                   & -                & -               & 20.2            & 35.6             & 409.5           & 563.7          & \textbf{1.00}         & 1.40          \\
			\multicolumn{1}{|l|}{SaCNN~\cite{deep-fscale-multitask-SaCNN-18}}           & WACV                  & 2018                   & 86.8             & 139.2           & 16.2            & 25.8             & 314.9           & 424.8          & -             & -             \\
			\multicolumn{1}{|l|}{ACSCP~\cite{deep-gan-loss-ACSCP-18}}           & CVPR                  & 2018                   & 75.7             & 102.7           & 17.2            & 27.4             & 291.0           & 404.6          & -             & -             \\
			\multicolumn{1}{|l|}{IG-CNN~\cite{deep-auto-IGCNN-18}}          & CVPR                  & 2018                   & 72.5             & 118.2           & 13.6            & 21.1             & 291.4           & 349.4          & -             & -             \\
			\multicolumn{1}{|l|}{Deep-NCL~\cite{deep-training-ConvNet-18}}        & CVPR                  & 2018                   & 73.5             & 112.3           & 18.7            & 26.0             & 288.4           & 404.7          & -             & -             \\ \hline
			\multicolumn{1}{|l|}{MCNN~\cite{MCNN-16}}            & CVPR                  & 2016                   & 110.2            & 173.2           & 26.4            & 41.3             & 377.6           & 509.1          & 1.07          & 1.35          \\
			\multicolumn{1}{|l|}{CSRNet~\cite{CSRNet-18}}          & CVPR                  & 2018                   & 68.2             & 115.0           & 10.6            & 16.0             & 266.1           & 397.5          & 1.16          & 1.47          \\
			\multicolumn{1}{|l|}{SANet~\cite{SANet-18}}           & ECCV                  & 2018                   & 67.0             & 104.5           & 8.4             & 13.6             & 258.4           & 334.9          & 1.02          & 1.29          \\ \hline
			\multicolumn{1}{|l|}{MCNN+SPANet}     & \multicolumn{1}{c}{-} & \multicolumn{1}{c|}{-} & 99.7             & 146.3            & 19.1             & 28.7              & 292.5            & 401.3           & \textbf{1.00}           & 1.33           \\
			\multicolumn{1}{|l|}{CSRNet+SPANet}   & \multicolumn{1}{c}{-} & \multicolumn{1}{c|}{-} & 62.4              & 99.5            & 8.4              & 13.2              & 245.8            & 333.1           & 1.12           & 1.42           \\
			\multicolumn{1}{|l|}{SANet+SPANet}    & \multicolumn{1}{c}{-} & \multicolumn{1}{c|}{-} & \textbf{59.4}     & \textbf{92.5}    & \textbf{6.5}     & \textbf{9.9}     & \textbf{232.6}   & \textbf{311.7}  & \textbf{1.00}  & \textbf{1.28} \\ \hline
		\end{tabular}
	\end{center}
	\vspace{-8mm}
\end{table*}

\section{Experiment}
\subsection{Experiment Settings}
\textbf{Networks}.~We evaluate our method by combining it with three networks, i.e., MCNN~\cite{MCNN-16}, CSRNet~\cite{CSRNet-18}, and SANet~\cite{SANet-18}. 
The implementations of MCNN\footnote{https://github.com/svishwa/crowdcount-mcnn} and CSRNet\footnote{https://github.com/leeyeehoo/CSRNet-pytorch/tree/master} are from others, while SANet is implemented by us. 
In general, there are four main differences between them: (1)~Different size of networks. Specifically, MCNN, SANet, and CSRNet are corresponding to small, medium, and large crowd counting networks. (2)~Different architectures. MCNN and SANet are multi-column/multi-blob networks, while CSRNet is a single column network. In addition, SANet uses the Instance Normalization (IN) layer and the deconvolutional layer, while CSRNet utilizes the dilated convolutional layer. (3)~Different size of density maps. Density maps of MCNN and CSRNet are 1/4 and 1/8 of original images, while SANet produces density maps with the same size as input images. (4)~Different testing scheme. SANet is tested on image patches, while CSRNet and MCNN are tested on the whole images. 

\textbf{Learning settings}.~For MCNN and SANet, the parameters are randomly initialized by a Gaussian distribution with mean of $0$ and standard deviation of $0.01$.
Adam optimizer~\cite{adam-14} with a learning rate of $1\mathrm{e}{-5}$ is used to train the model.
For CSRNet, the first ten convolutional layers are from pre-trained VGG-16~\cite{vgg-14}.
The other layers are initialized in the same way as MCNN.
Stochastic gradient descent (SGD) with a fixed learning rate of $1\mathrm{e}{-6}$ is applied during the training.

\textbf{Datasets}.~We evaluate our method on four datasets, including ShanghaiTech~\cite{MCNN-16},~UCF\_CC\_50~\cite{r-mrf-count-multi-source-13},~WorldExpo'10~\cite{deep-crowd-scene-15},~and UCSD~\cite{r-count-privacy-preserving-08}. Typically, ShanghaiTech Part A is congested and noisy, while ShanghaiTech Part B is noisy but not highly congested. UCF\_CC\_50 consists of extremely congested scenes with heavy background noises. WorldExpo'10 and UCSD contain sparse crowd scenes. The scenes in WorldExpo'10 are noisier than UCSD.

\textbf{Evaluation details}.~MCNN and CSRNet are tested on the whole images, while SANet is tested on image patches. 
Following previous works~\cite{CSRNet-18,deep-training-ICCNN-18,MCNN-16}, Mean Absolute Error (MAE) and Mean Square Error (MSE) are used to evaluate the performance by
\begin{equation}
\footnotesize
\label{eq:mae-mse}
MAE=\frac{1}{N}\sum_{i=1}^N\left | C_i-C_i^{gt} \right |,\;\;
MSE=\sqrt{\frac{1}{N}\sum_{i=1}^N\left ( C_i-C_i^{gt} \right )^2},
\end{equation}
where $C_i$ is the estimated crowd count and $C_i^{gt}$ is the ground truth count of the $i$-th image. 
$N$ is the number of test images. 
Additionally, PSNR (Peak Signal-to-Noise Ratio)\footnote{https://en.wikipedia.org/wiki/Peak\_signal-to-noise\_ratio} and SSIM (Structural Similarity)\footnote{https://en.wikipedia.org/wiki/Structural\_similarity}~\cite{ssim} are utilized to measure the quality of density maps. 
For fair comparison, similar to \cite{CSRNet-18}, bilinear interpolation is employed to resize estimated density maps to the same size as input images.

\subsection{Comparisons with State-of-the-art}
Table \ref{tab:stoa-1} and \ref{tab:sota-2} report the results of four challenging datasets.
As a summary, our method significantly improves all baselines and outperforms the other state-of-the-art methods.
This result fully demonstrates the effectiveness of our SPANet, which could provide accurate density estimation on both dense and sparse crowd scenes, and can be applied to all CNN-based crowd counting networks.

On ShanghaiTech dataset, our SPANet boosts MCNN, CSRNet, SANet with relative MAE improvements of 9.5\%, 8.5\%, 11.3\% on Part A, and 27.7\%, 20.8\%, 22.7\% on Part B.
Noted that Part A is collected from the internet while Part B is from the busy streets and has more spatial constraints.
Since our SPANet can fully utilize spatial awareness, it brings more improvements on Part B.
On UCF\_CC\_50, SPANet provides the relative MAE improvements of 22.5\%, 7.6\%, 10.0\% for the three baselines.
Noted that the improved MCNN is even comparable with other state-of-the-art methods.
It clearly shows that SPANet can handle the extremely dense-crowd scenes.
Similar to the above two datasets, SPANet also achieves significant improvements on UCSD and WorldExpo'10, verifying the effectiveness of our method on the sparse-crowd scenes.

\begin{table}[!t]
	\footnotesize
	\begin{center}
		\caption{\footnotesize Comparison with the state-of-the-art methods on WorldExpo'10~\cite{r-count-privacy-preserving-08} dataset. Only MAE is computed for each scene and then averaged to evaluate the overall performance.}
		\label{tab:sota-2}
		\vspace{-0.13in}
		\begin{tabular}{|l|c|c|c|c|c|c|}
			\hline
			\textbf{Method}        & \textbf{S1} & \textbf{S2} & \textbf{S3} & \textbf{S4} & \textbf{S5} & \textbf{Avg.} \\ \hline \hline
			Zhang et al.~\cite{deep-crowd-scene-15}  & 9.8    & 14.1   & 14.3   & 22.2   & 3.7    & 12.9     \\
			Huang et al.~\cite{deep-body-BSAD-18}  & 4.1    & 21.7   & 11.9   & 11.0   & 3.5    & 10.5     \\
			Switch-CNN~\cite{deep-switch-17}    & 4.4    & 15.7   & 10.0   & 11.0   & 5.9    & 9.4      \\
			SaCNN~\cite{deep-fscale-multitask-SaCNN-18}   &2.6    &13.5    &10.6    &12.5   &3.3    &8.5 \\
			CP-CNN~\cite{deep-multi-pyramid-cnn-17}        & 2.9    & 14.7   & 10.5   & \textbf{10.4}   & 5.8    & 8.9      \\ \hline
			MCNN~\cite{MCNN-16}          & 3.4    & 20.6   & 12.9   & 13.0   & 8.1    & 11.6     \\ 
			CSRNet~\cite{CSRNet-18}        & 2.9    & {11.5}   & 8.6    & 16.6   & 3.4    & 8.6      \\ 
			SANet~\cite{SANet-18}         & 2.6    & 13.2   & 9.0    & 13.3   & \textbf{3.0}    & 8.2      \\ \hline
			MCNN+SPANet    & 3.4     & 14.9    & 15.1     & 12.8    & 4.5     & 10.1      \\
			CSRNet+SPANet & 2.6     & \textbf{11.1}    & 8.9      & 13.5    & 3.3     & 7.9       \\ 
			SANet+SPANet   & \textbf{2.3}     & 12.3    & \textbf{7.9}     & 12.9    & 3.2      & \textbf{7.7}       \\ \hline
		\end{tabular}
	\end{center}
	\vspace{-6mm}
\end{table}

\begin{figure}[]
	\footnotesize
	\vspace{-1mm}
	\centering
	\subfloat{{
			\centering
			\includegraphics[width=0.48\linewidth]{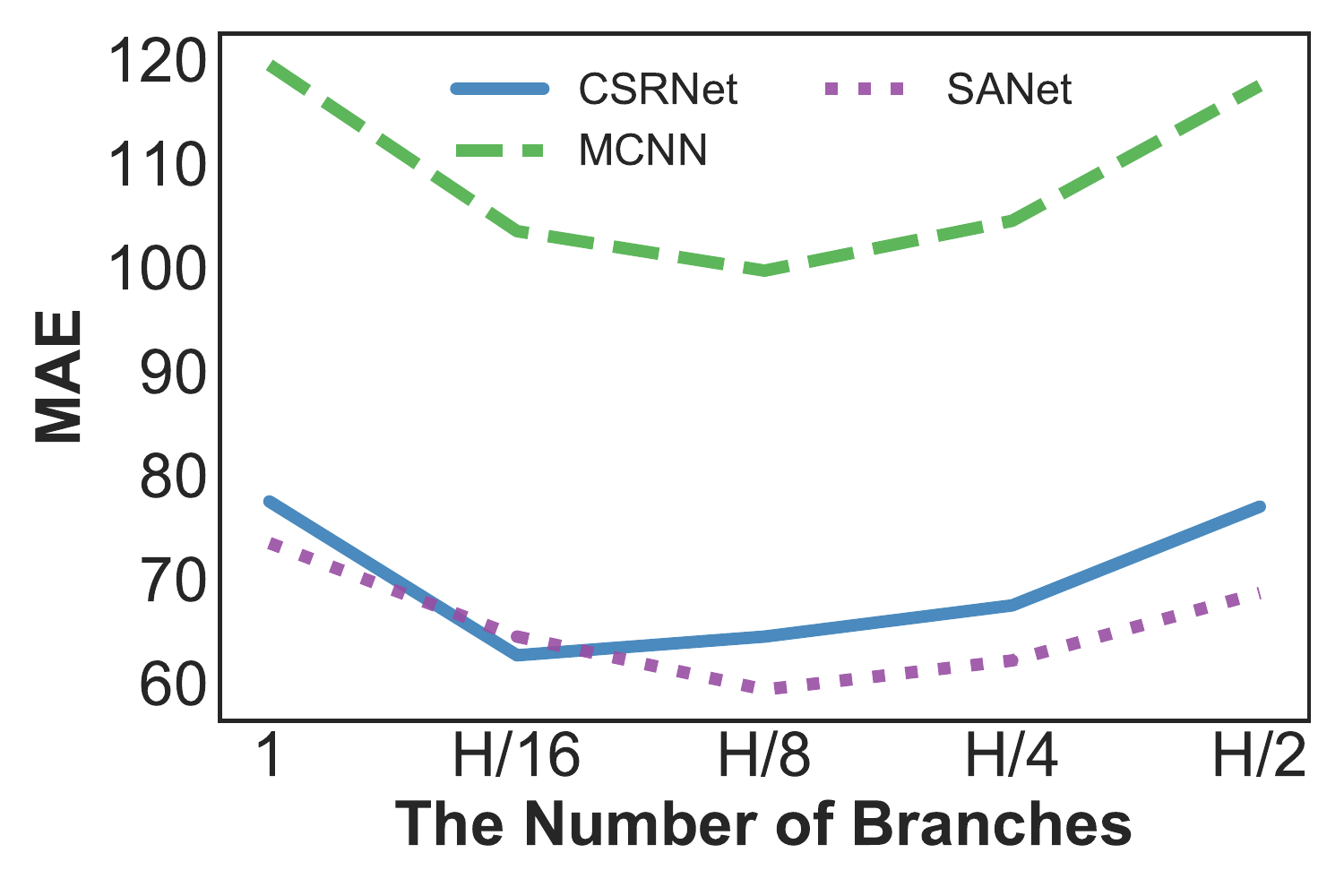}}}%
	\subfloat{{
			\centering
			\includegraphics[width=0.48\linewidth]{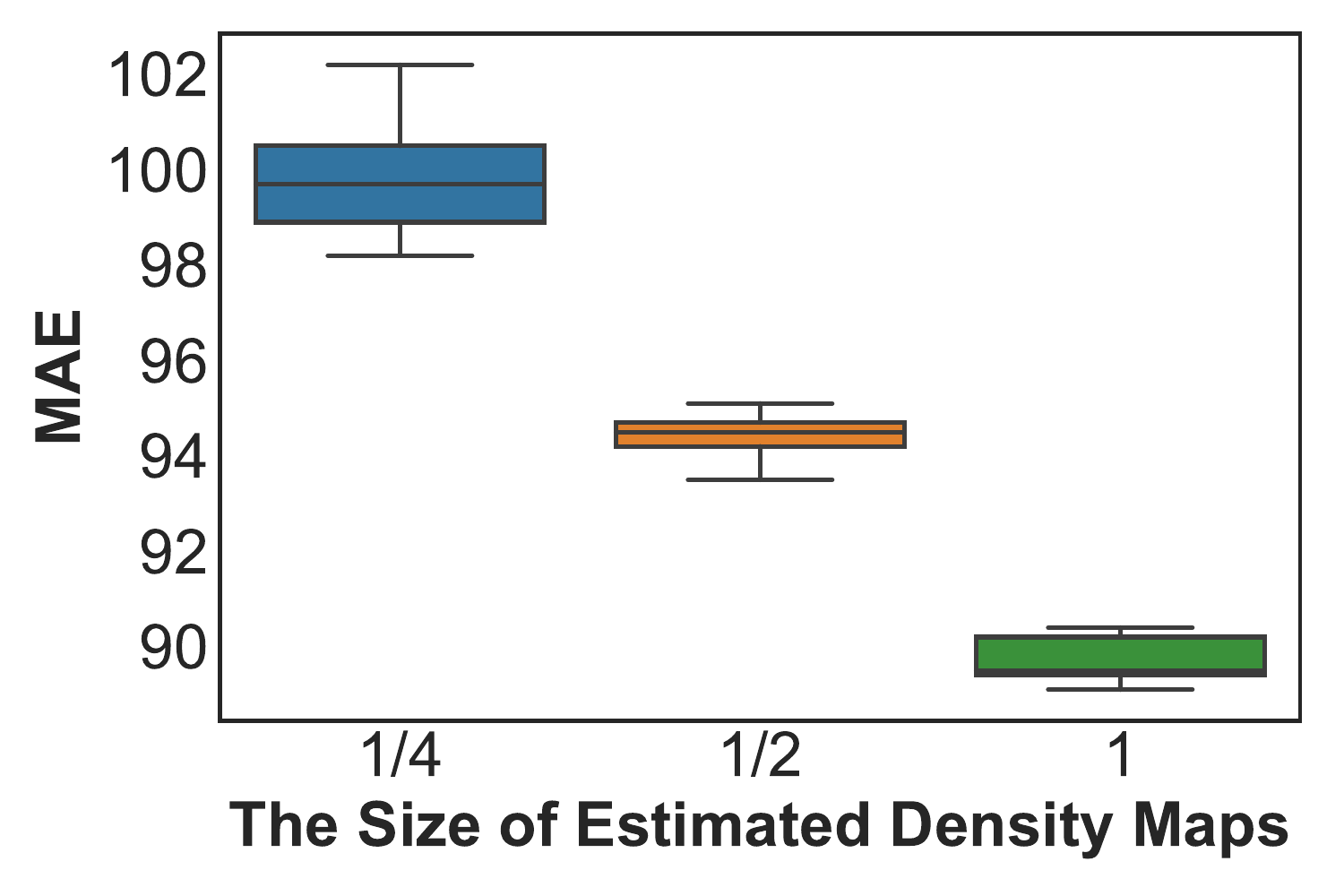} }}%
	\vspace{-2mm}
	\caption{\footnotesize Ablation studies on ShanghaiTech Part A \cite{MCNN-16}. The left shows the branch number $K$ vs. MAE, and the right illustrates the size of estimated density maps vs. MAE, performed with MCNN.
	} 
	\label{fig:ablation-study}
	\vspace{-4mm}
\end{figure}

\begin{figure*}[!thb]
	\footnotesize
	\centering
	\includegraphics[width=0.88\textwidth]{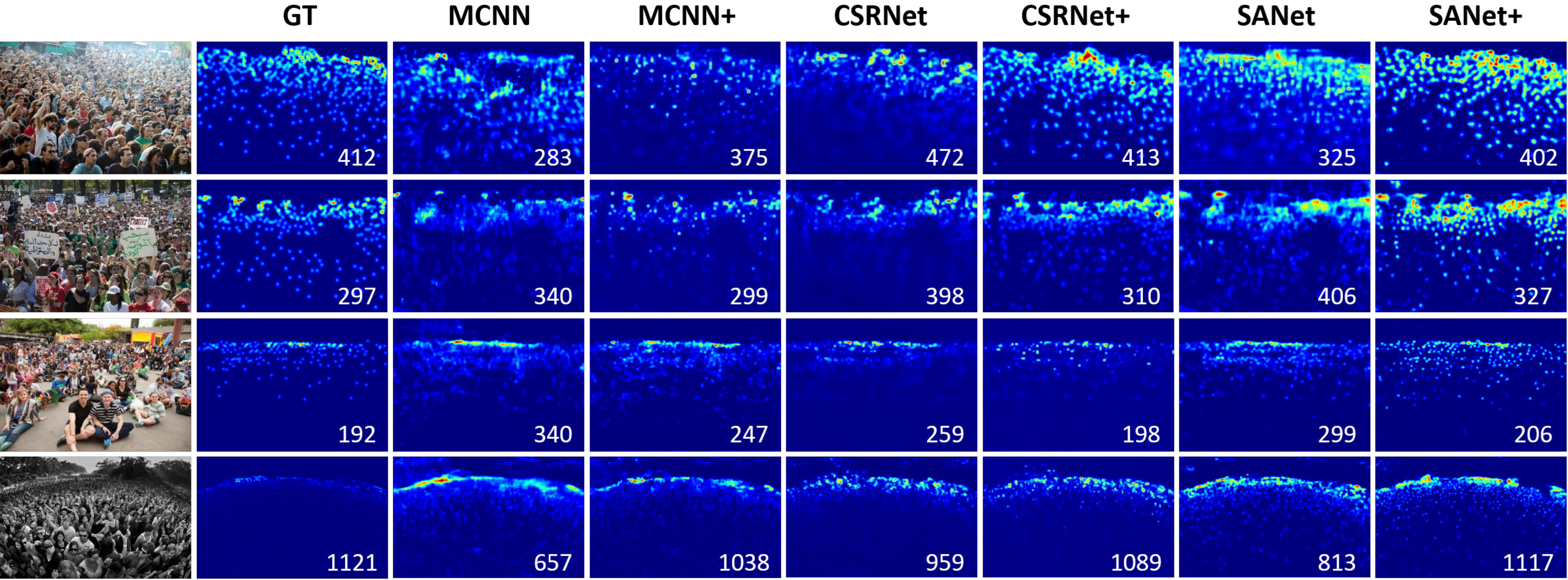}
	\vspace{-2mm}
	\caption{\footnotesize Comparisons of estimated density maps between baselines and our SPANet. `+' indicates combining SPANet with baselines.}
	\label{fig:qualitative}
	\vspace{-2mm}
\end{figure*}

\begin{figure*}[!th]
	\vspace{-3mm}
	\footnotesize
	\centering
	\subfloat{{
			\centering
			\includegraphics[width=0.27\linewidth]{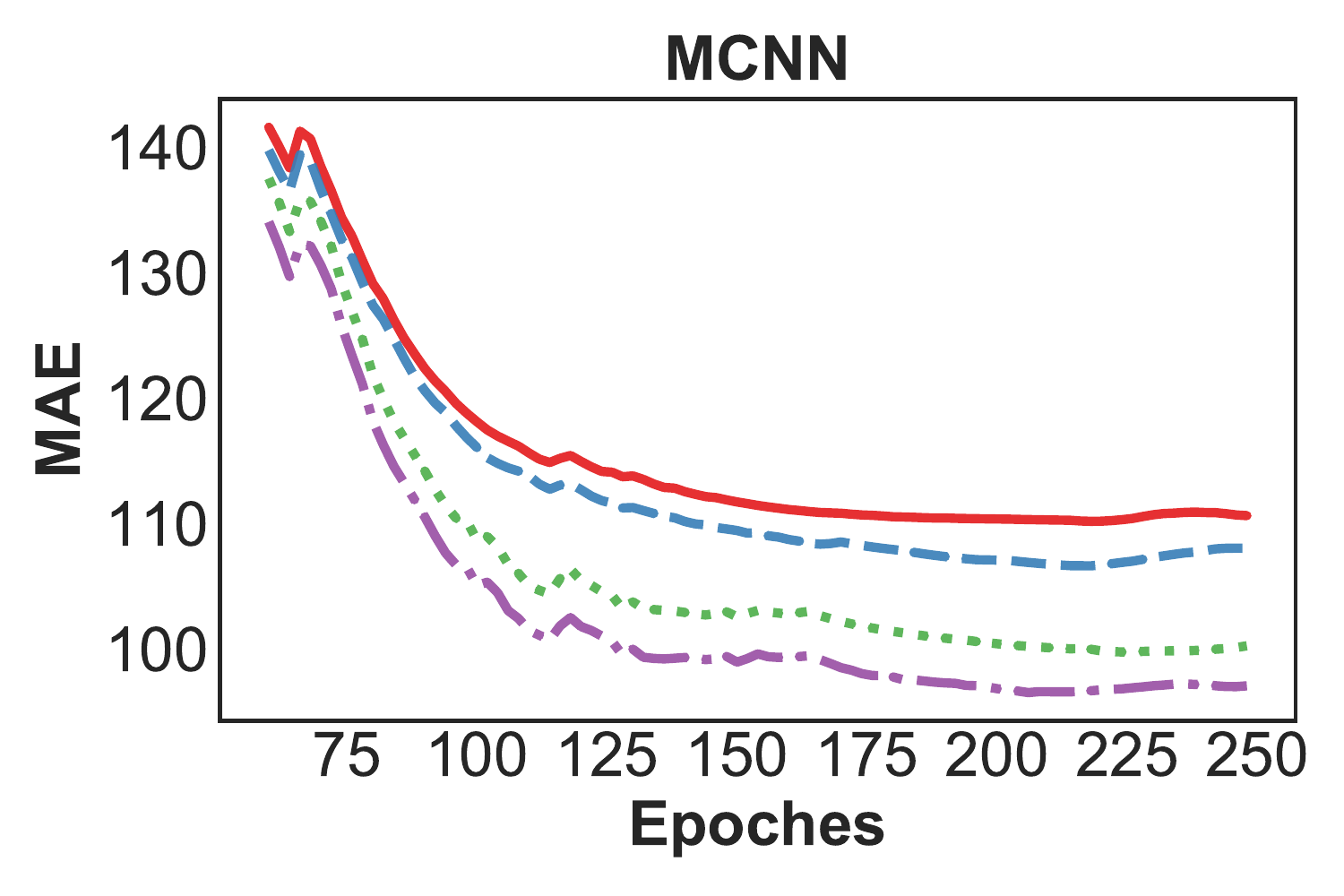}}}%
	\qquad
	\subfloat{{
			\centering
			\includegraphics[width=0.27\linewidth]{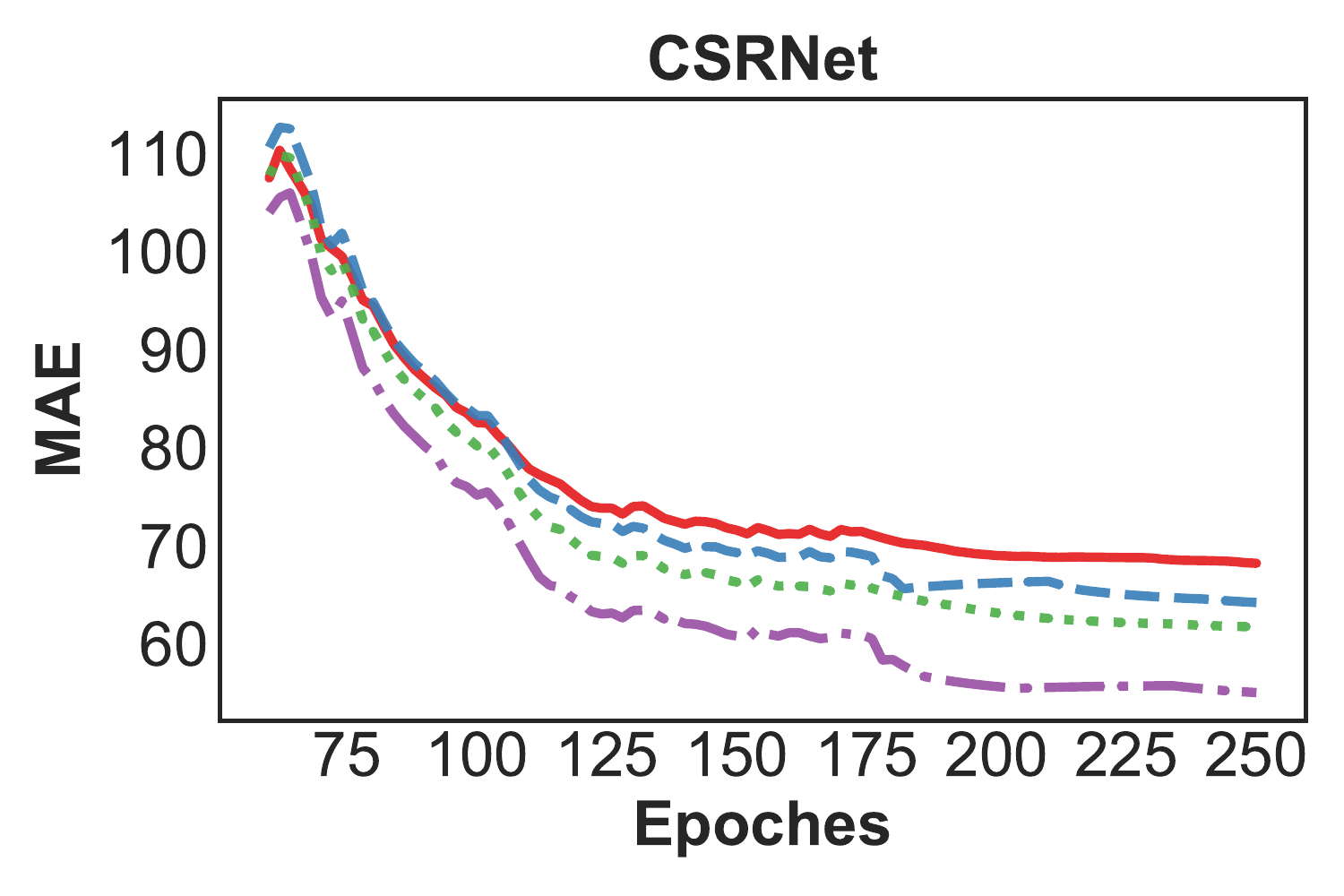} }}%
	\qquad
	\subfloat{{
			\centering
			\includegraphics[width=0.27\linewidth]{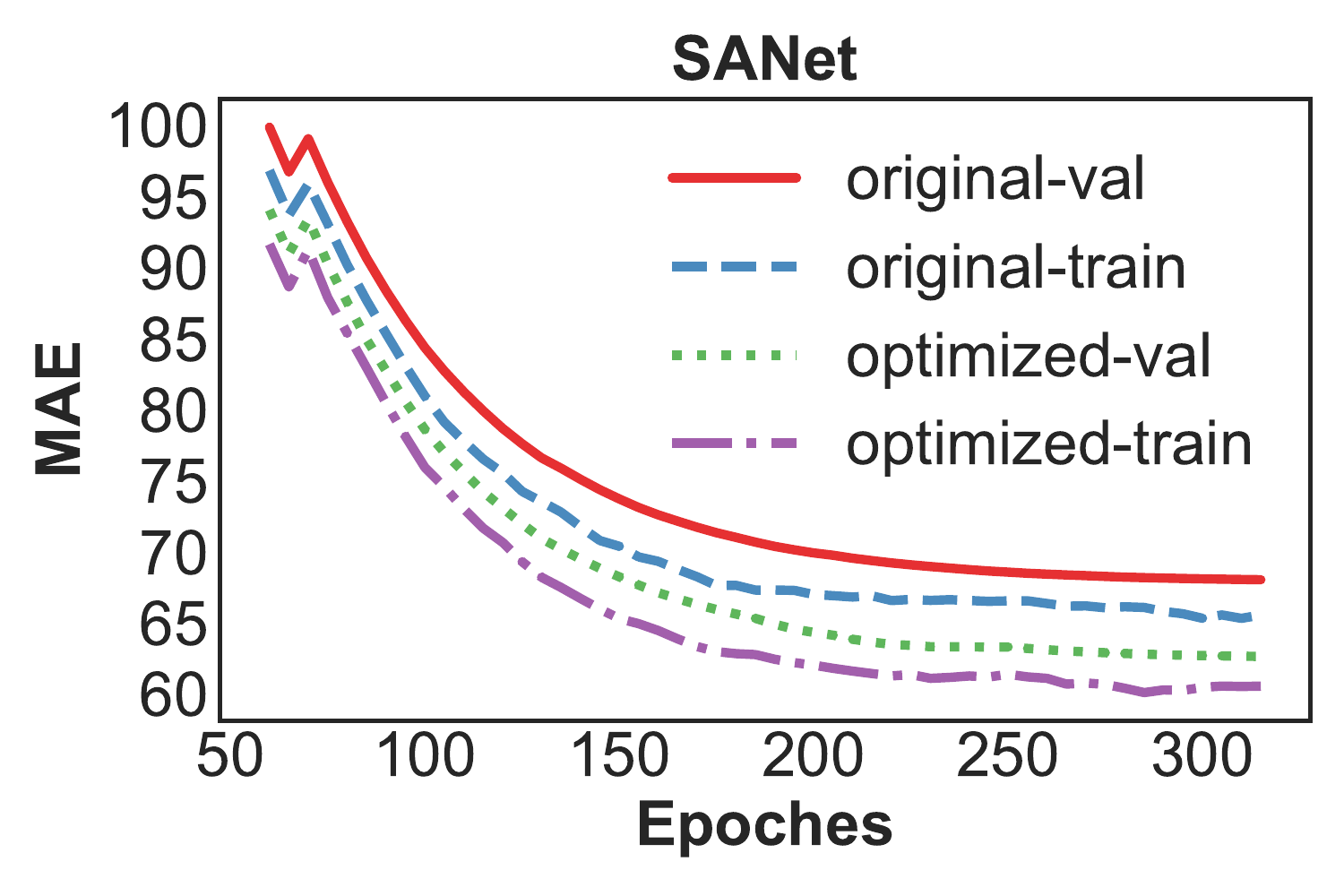} }}%
	\vspace{-3mm}
	\caption{\footnotesize Learning Curves. Mean absolute error (MAE) on training and validation sets, vs. the number of training epochs of MCNN \cite{MCNN-16}, CSRNet~\cite{CSRNet-18} and SANet~\cite{SANet-18} on ShanghaiTech Part A dataset~\cite{MCNN-16}.} 
	\label{fig:learning-curve}
	\vspace{-1mm}
\end{figure*}

\subsection{Ablation Studies}

\noindent\textbf{Sampling positions}.~We first evaluate the impact of different starting positions when sampling patches for mask pooling.
The results are listed in Table~\ref{tab:ablation-study}.
We find that starting at the bottom is always better than the top, and the right is also better than the left.
The possible reason is that it may be closely related to camera calibration.
The results encourage us to sample patches from the bottom right corner.
Noted that the differences between these sampling schemes are quite small, which demonstrates the robustness of our method.
Additionally, we also present the comparison of performing mask pooling on the same or different density maps in each branch, which is already discussed in Section \ref{sec:optimized-l-mesa} and Eq. (\ref{eqn:crowd-count}).
As shown in Table~\ref{tab:ablation-study}, the results of two strategies are similar.
Due to the efficiency problem, we directly pool patches from the same density map.


\begin{table}[!t]
	\footnotesize
	\begin{center}
		\caption{\footnotesize Ablation studies of patch sampling strategy, mask pooling strategy, and losses on ShanghaiTech Part A dataset~\cite{MCNN-16}.}
		\label{tab:ablation-study}
		\vspace{-0.1in}
		\begin{tabular}{|l|r|r|} 
			\hline
			\textbf{Configurations} & \multicolumn{1}{l|}{\textbf{MAE} $\downarrow$} & \multicolumn{1}{l|}{\textbf{MSE} $\downarrow$}  \\ 
			\hline \hline
			Center point                 & 101.2                             & 153.3                               \\
			Top left corner                & 101.5                             & 153.7                               \\
			Bottom left corner             & 100.7                             & 149.2                               \\
			Top right corner              & 100.5                             & 149.4                               \\ 
			Bottom right corner           & 99.7                      & 146.3                   \\ \hline \hline
			Different density map            & 100.3                             & 147.4                       \\ 
			Same density map                 & \textbf{99.7}                      & \textbf{146.3}            \\ \hline \hline
			$L_2$                    & 110.2                             & 173.2                               \\
			$L_r$+$L_{mep}$ & 99.3                            & 145.3                               \\
			$L_2+L_r$               & 107.2                             & 164.5                               \\
			$L_2+L_r+L_{mep}$    & 99.7                          & 146.3                               \\
		      Random & 105.4 & 162.2                             \\
		Grid Search & \textbf{98.3} & \textbf{142.5}                        \\
			\hline
		\end{tabular}
	\end{center}
	\vspace{-8mm}
\end{table}

\begin{table*}[!t]
	\footnotesize
	\begin{center}
		\caption{\footnotesize Density map quality comparison. Values on the left of `$\mid$' are from original baselines, while values on the right of `$\mid$' are results when integrating with the proposed SPANet.}
		\label{tab:quality-density-map}
		\vspace{-0.1in}
		\begin{tabular}{|l|c|c|c|c|c|c|}
			\hline
			& \multicolumn{2}{c|}{\textbf{MCNN}} & \multicolumn{2}{c|}{\textbf{CSRNet}} & \multicolumn{2}{c|}{\textbf{SANet}}\\ \hline
			\textbf{Dataset} & \textbf{PSNR}  $\uparrow$  & \textbf{SSIM}  $\uparrow$  & \textbf{PSNR}  $\uparrow$  & \textbf{SSIM}  $\uparrow$  & \textbf{PSNR}  $\uparrow$  & \textbf{SSIM}  $\uparrow$   \\ \hline \hline
			ShanghaiTech-A~\cite{MCNN-16} & 21.42 $\mid$ 22.18 & 0.52 $\mid$ 0.66 & 23.79 $\mid$ 24.88 & 0.76 $\mid$ 0.85 & 23.36 $\mid$ 25.33 & 0.78 $\mid$ 0.85 \\
			ShanghaiTech-B~\cite{MCNN-16} & 23.43 $\mid$ 26.19 & 0.78 $\mid$ 0.85 & 27.02 $\mid$ 29.50 & 0.89 $\mid$ 0.92 & 27.44 $\mid$ 29.17 & 0.89 $\mid$ 0.91 \\
			UCF\_CC\_50~\cite{r-mrf-count-multi-source-13}         & 14.44 $\mid$ 18.25 & 0.37 $\mid$ 0.51 & 18.76 $\mid$ 20.17 & 0.52 $\mid$ 0.78 & 18.35 $\mid$ 20.01 & 0.51 $\mid$ 0.76 \\
			UCSD~\cite{deep-crowd-scene-15}                 & 17.43 $\mid$ 18.52 & 0.75 $\mid$ 0.83 & 20.02 $\mid$ 21.80 & 0.86 $\mid$ 0.89 & 21.33 $\mid$ 22.20 & 0.84 $\mid$ 0.90\\ 
			WorldExpo'10~\cite{r-count-privacy-preserving-08}        & 23.53 $\mid$ 25.97 & 0.76 $\mid$ 0.85 & 26.94 $\mid$ 29.05 & 0.92 $\mid$ 0.93 & 26.22 $\mid$ 28.54 & 0.90 $\mid$ 0.92 \\
			\hline
		\end{tabular}
	\end{center}
	\vspace{-8mm}
\end{table*}


\noindent\textbf{Different losses/weights}.~We turn to evaluate the effect of different losses and weight schemes.
As shown in Table \ref{tab:ablation-study}, adding the ranking loss only provides slight improvement, while the significant improvement comes from the MEP loss.
Besides, there is no significant difference whether $L_2$ is used.
It demonstrates that our MEP loss can effectively learn spatial awareness to boost crowd counting.
We further conduct experiments on two weight schemes: the random weight and the grid search with step 0.1. 
As shown in Table \ref{tab:ablation-study}, our method is not sensitive to the weights. Even the grid search brings a very slight improvement.

\noindent\textbf{Number of branches}.~We measure the performance of SPANet with different branch numbers ${\small K}$.
As illustrated in Figure \ref{fig:ablation-study}, the performance first improves but then drops with the increasing number of ${\small K}$.
This observation is not surprising. 
On one side, small ${\small K}$ (e.g., ${\small K=1}$) would involve large bias error due to the large difference between two patches.
On the other side, large ${\small K}$ (e.g., ${\small K=\frac{{H}}{2}}$, where ${\small H}$ is the height of estimated density map) implies that the difference of two patches in each branch is very small, which cannot provide enough discrepancy for subregion generation.
In experiments, $K$ is set to ${\small \frac{{H}}{8}}$ for MCNN/SANet and ${\small \frac{{H}}{16}}$ for CSRNet, which is determined via cross validation.

\noindent\textbf{Size of estimated density maps}.~We further validate the effect of the size of estimated density maps.
We add deconvolutional layers on top of the MCNN to increase the size of the estimated density maps. 
Eventually, two variants of MCNN are obtained, whose estimated density maps are of $1/2$ and the same size as the input images, respectively.
As shown in Figure \ref{fig:ablation-study}, the performance is improved along with the size increase of density maps.
The results indicate that predicting high-resolution density maps could bring considerable improvement.


\subsection{Studies on Estimated Density Maps}
We now evaluate the estimated density maps to verify whether our method can fully utilize spatial awareness.
Table~\ref{tab:quality-density-map} summarizes the results.
Our SPANet can significantly improve PSNR and SSIM across all baselines and datasets, which indicates that the quality of the generated density maps are significantly improved.
To further verify that our method can indeed learn spatial awareness, we showcase the generated density maps of four examples from different methods in Figure~\ref{fig:qualitative}.
These four examples typically contain different crowd densities, occlusions, and scale changes.
We can observe that the baseline models are always affected by the zero-mean noise, which leads to overestimation in low-density areas.
In contrast, zero-mean noise is effectively suppressed in our SPANet. 
Besides, baseline models normally have an insufficient estimation for high-density areas, while ours can obtain a more accurate estimation for them.
Noted that the ground truth itself is also generated with center points of pedestrians' heads, which inherently contains inaccurate information.
It means that our method is still unable to produce the same density map to the ground truth.

\subsection{Studies on Learning Curves}
Finally, we study the learning curves to further evaluate our method. 
Figure~\ref{fig:learning-curve} shows the training and validation mean absolute error (MAE) at every epoch on ShanghaiTech Part A dataset.
For better viewing, we smooth the learning curves by exponential moving average (EMA) with a smoothing factor $\alpha = 0.1$. 
Compared with original results, baselines integrated with our SPANet exhibit lower MAE on both training and testing set. 
Since the performance on the training and testing set generally denotes the fitting and generalization degree, this result demonstrates the promising capability on both sides.
In addition, it also means that our method can significantly improve the stability during model training.

\vspace{-2mm}
\section{Conclusion}
\vspace{-1mm}
\label{sec:conclusion}
In this paper we present a novel deep architecture called SPatial Awareness Network (SPANet) for crowd counting, which is able to capture the spatial variations by finding the pixel-level subregion with high discrepancy to the ground truth.
It could be integrated into all CNN-based methods and is end-to-end trainable.
Experiments on four datasets and three various networks fully demonstrate that it can significantly improve all baselines and outperforms the state-of-the-art methods.
It provides the elegant views of effectively using spatial awareness to improve crowd counting.
In future work we will study how to preserve spatial awareness as much as possible in the ground truth generation.

\vspace{-2mm}
\section*{Acknowledgements}
\vspace{-2mm}
{\small This research was supported in part through the financial assistance award 60NANB17D156 from U.S. Department of Commerce, National Institute of Standards and Technology and by the Intelligence Advanced Research Projects Activity (IARPA) via Department of Interior/Interior Business Center (DOI/IBC) contract number D17PC00340, National Natural Science Foundation of China (61772436), Foundation for Department of Transportation of Henan Province, China (2019J-2-2), Sichuan Science and Technology Innovation Seedling Fund (2017RZ0015), China Scholarship Council (201707000083) and Cultivation Program for the Excellent Doctoral Dissertation of Southwest Jiaotong University (D-YB 201707).}

{\small
\bibliographystyle{ieee_fullname}
\bibliography{ms}

\begin{thebibliography}{10}\itemsep=-1pt

\bibitem{deep-shallow-deep-crowdnet-16}
Lokesh Boominathan, Srinivas S.~S. Kruthiventi, and R.~Venkatesh Babu.
\newblock Crowdnet: {A} deep convolutional network for dense crowd counting.
\newblock In {\em Proceedings of ACM International Conference on Multimedia},
  pages 640--644, 2016.

\bibitem{d-bayesian-full-06}
Gabriel~J Brostow and Roberto Cipolla.
\newblock Unsupervised bayesian detection of independent motion in crowds.
\newblock In {\em Proceedings of IEEE Conference on Computer Vision and Pattern
  Recognition}, volume~1, pages 594--601, 2006.

\bibitem{SANet-18}
Xinkun Cao, Zhipeng Wang, Yanyun Zhao, and Fei Su.
\newblock Scale aggregation network for accurate and efficient crowd counting.
\newblock In {\em Proceedings of European Conference on Computer Vision}, pages
  757--773, 2018.

\bibitem{r-count-privacy-preserving-08}
Antoni~B Chan, Zhang-Sheng~John Liang, and Nuno Vasconcelos.
\newblock Privacy preserving crowd monitoring: Counting people without people
  models or tracking.
\newblock In {\em Proceedings of IEEE Conference on Computer Vision and Pattern
  Recognition}, pages 1--7, 2008.

\bibitem{r-edge-count-bayesian-12}
Antoni~B Chan and Nuno Vasconcelos.
\newblock Counting people with low-level features and bayesian regression.
\newblock {\em IEEE Transactions on Image Processing}, 21(4):2160--2177, 2012.

\bibitem{MM-2018}
Zhi-Qi Cheng, Jun-Xiu Li, Qi Dai, Xiao Wu, Jun-Yan He, and Alexander Hauptmann.
\newblock Improving the learning of multi-column convolutional neural network
  for crowd counting.
\newblock In {\em Proceedings of the 26th ACM International Conference on
  Multimedia}, 2019.

\bibitem{d-histograms-svm-full-05}
Navneet Dalal and Bill Triggs.
\newblock Histograms of oriented gradients for human detection.
\newblock In {\em Proceedings of IEEE Conference on Computer Vision and Pattern
  Recognition}, volume~1, pages 886--893, 2005.

\bibitem{d-part2-08}
Piotr Doll{\'a}r, Boris Babenko, Serge Belongie, Pietro Perona, and Zhuowen Tu.
\newblock Multiple component learning for object detection.
\newblock In {\em Proceedings of European Conference on Computer Vision}, pages
  211--224, 2008.

\bibitem{SPooling-18}
Siyu Huang, Xi Li, Zhiqi Cheng, Zhongfei Zhang, and Alexander~G. Hauptmann.
\newblock Stacked pooling: Improving crowd counting by boosting scale
  invariance.
\newblock {\em CoRR}, abs/1808.07456, 2018.

\bibitem{deep-body-BSAD-18}
Siyu Huang, Xi Li, Zhongfei Zhang, Fei Wu, Shenghua Gao, Rongrong Ji, and
  Junwei Han.
\newblock Body structure aware deep crowd counting.
\newblock {\em IEEE Transactions on Image Processing}, 27(3):1049--1059, 2018.

\bibitem{r-mrf-count-multi-source-13}
Haroon Idrees, Imran Saleemi, Cody Seibert, and Mubarak Shah.
\newblock Multi-source multi-scale counting in extremely dense crowd images.
\newblock In {\em Proceedings of IEEE Conference on Computer Vision and Pattern
  Recognition}, pages 2547--2554, 2013.

\bibitem{d-part-15}
Haroon Idrees, Khurram Soomro, and Mubarak Shah.
\newblock Detecting humans in dense crowds using locally-consistent scale prior
  and global occlusion reasoning.
\newblock {\em IEEE Transactions on Pattern Analysis and Machine Intelligence},
  37(10):1986--1998, 2015.

\bibitem{deep-multi-attention-AFP-18}
Di Kang and Antoni~B. Chan.
\newblock Crowd counting by adaptively fusing predictions from an image
  pyramid.
\newblock In {\em Proceedings of British Machine Vision Conference}, page~89,
  2018.

\bibitem{adam-14}
Diederik~P Kingma and Jimmy Ba.
\newblock Adam: A method for stochastic optimization.
\newblock {\em arXiv preprint arXiv:1412.6980}, 2014.

\bibitem{ledig2017photo}
Christian Ledig, Lucas Theis, Ferenc Husz{\'a}r, Jose Caballero, Andrew
  Cunningham, Alejandro Acosta, Andrew Aitken, Alykhan Tejani, Johannes Totz,
  Zehan Wang, et~al.
\newblock Photo-realistic single image super-resolution using a generative
  adversarial network.
\newblock In {\em Proceedings of IEEE Conference on Computer Vision and Pattern
  Recognition}, pages 4681--4690, 2017.

\bibitem{nips-10}
Victor~S. Lempitsky and Andrew Zisserman.
\newblock Learning to count objects in images.
\newblock In {\em Proceedings of Conference on Neural Information Processing
  Systems}, pages 1324--1332, 2010.

\bibitem{CSRNet-18}
Yuhong Li, Xiaofan Zhang, and Deming Chen.
\newblock Csrnet: Dilated convolutional neural networks for understanding the
  highly congested scenes.
\newblock In {\em Proceedings of IEEE Conference on Computer Vision and Pattern
  Recognition}, pages 1091--1100, 2018.

\bibitem{d-harr-svm-part-01}
Sheng-Fuu Lin, Jaw-Yeh Chen, and Hung-Xin Chao.
\newblock Estimation of number of people in crowded scenes using perspective
  transformation.
\newblock {\em IEEE Transactions on Systems, Man, and Cybernetics-Part A:
  Systems and Humans}, 31(6):645--654, 2001.

\bibitem{deep-decide-18}
Jiang Liu, Chenqiang Gao, Deyu Meng, and Alexander~G. Hauptmann.
\newblock Decidenet: Counting varying density crowds through attention guided
  detection and density estimation.
\newblock In {\em Proceedings of IEEE Conference on Computer Vision and Pattern
  Recognition}, pages 5197--5206, 2018.

\bibitem{deep-attention-cnn-ADCrowdNet-18}
Ning Liu, Yongchao Long, Changqing Zou, Qun Niu, Li Pan, and Hefeng Wu.
\newblock Adcrowdnet: An attention-injective deformable convolutional network
  for crowd understanding.
\newblock {\em CoRR}, abs/1811.11968, 2018.

\bibitem{deep-Geometric-18}
Weizhe Liu, Krzysztof Lis, Mathieu Salzmann, and Pascal Fua.
\newblock Geometric and physical constraints for head plane crowd density
  estimation in videos.
\newblock {\em CoRR}, abs/1803.08805, 2018.

\bibitem{deep-fscale-Context-18}
Weizhe Liu, Mathieu Salzmann, and Pascal Fua.
\newblock Context-aware crowd counting.
\newblock {\em CoRR}, abs/1811.10452, 2018.

\bibitem{L2Rank-18}
Xialei Liu, Joost van~de Weijer, and Andrew~D. Bagdanov.
\newblock Leveraging unlabeled data for crowd counting by learning to rank.
\newblock In {\em Proceedings of IEEE Conference on Computer Vision and Pattern
  Recognition}, pages 7661--7669, 2018.

\bibitem{kolmogorov-51}
Frank~J Massey~Jr.
\newblock The kolmogorov-smirnov test for goodness of fit.
\newblock {\em Journal of the American statistical Association},
  46(253):68--78, 1951.

\bibitem{deep-multi-perspetive-free-16}
Daniel O{\~{n}}oro{-}Rubio and Roberto~Javier L{\'{o}}pez{-}Sastre.
\newblock Towards perspective-free object counting with deep learning.
\newblock In {\em Proceedings of European Conference on Computer Vision}, pages
  615--629, 2016.

\bibitem{r-forest-density-15}
Viet{-}Quoc Pham, Tatsuo Kozakaya, Osamu Yamaguchi, and Ryuzo Okada.
\newblock {COUNT} forest: Co-voting uncertain number of targets using random
  forest for crowd density estimation.
\newblock In {\em Proceedings of International Conference on Computer Vision},
  pages 3253--3261, 2015.

\bibitem{deep-training-ICCNN-18}
Viresh Ranjan, Hieu Le, and Minh Hoai.
\newblock Iterative crowd counting.
\newblock In {\em Proceedings of European Conference on Computer Vision}, pages
  278--293, 2018.

\bibitem{r-edge-count-96}
Carlo~S Regazzoni and Alessandra Tesei.
\newblock Distributed data fusion for real-time crowding estimation.
\newblock {\em Signal Processing}, 53(1):47--63, 1996.

\bibitem{r-density-edge-09}
David Ryan, Simon Denman, Clinton Fookes, and Sridha Sridharan.
\newblock Crowd counting using multiple local features.
\newblock In {\em Digital Image Computing: Techniques and Applications}, pages
  81--88, 2009.

\bibitem{deep-auto-TDF-CNN-18}
Deepak~Babu Sam and R.~Venkatesh Babu.
\newblock Top-down feedback for crowd counting convolutional neural network.
\newblock In {\em Proceedings of Conference on Artificial Intelligence}, pages
  7323--7330, 2018.

\bibitem{deep-auto-IGCNN-18}
Deepak~Babu Sam, Neeraj~N. Sajjan, R.~Venkatesh Babu, and Mukundhan Srinivasan.
\newblock Divide and grow: Capturing huge diversity in crowd images with
  incrementally growing {CNN}.
\newblock In {\em Proceedings of IEEE Conference on Computer Vision and Pattern
  Recognition}, pages 3618--3626, 2018.

\bibitem{deep-switch-17}
Deepak~Babu Sam, Shiv Surya, and R.~Venkatesh Babu.
\newblock Switching convolutional neural network for crowd counting.
\newblock In {\em Proceedings of IEEE Conference on Computer Vision and Pattern
  Recognition}, pages 4031--4039, 2017.

\bibitem{grad-17}
Ramprasaath~R Selvaraju, Michael Cogswell, Abhishek Das, Ramakrishna Vedantam,
  Devi Parikh, and Dhruv Batra.
\newblock Grad-cam: Visual explanations from deep networks via gradient-based
  localization.
\newblock In {\em Proceedings of the IEEE International Conference on Computer
  Vision}, pages 618--626, 2017.

\bibitem{deep-gan-loss-ACSCP-18}
Zan Shen, Yi Xu, Bingbing Ni, Minsi Wang, Jianguo Hu, and Xiaokang Yang.
\newblock Crowd counting via adversarial cross-scale consistency pursuit.
\newblock In {\em Proceedings of IEEE Conference on Computer Vision and Pattern
  Recognition}, pages 5245--5254, 2018.

\bibitem{Perspective-18}
Miaojing Shi, Zhaohui Yang, Chao Xu, and Qijun Chen.
\newblock Perspective-aware {CNN} for crowd counting.
\newblock {\em CoRR}, abs/1807.01989, 2018.

\bibitem{deep-training-ConvNet-18}
Zenglin Shi, Le Zhang, Yun Liu, Xiaofeng Cao, Yangdong Ye, Ming{-}Ming Cheng,
  and Guoyan Zheng.
\newblock Crowd counting with deep negative correlation learning.
\newblock In {\em Proceedings of IEEE Conference on Computer Vision and Pattern
  Recognition}, pages 5382--5390, 2018.

\bibitem{vgg-14}
Karen Simonyan and Andrew Zisserman.
\newblock Very deep convolutional networks for large-scale image recognition.
\newblock {\em arXiv preprint arXiv:1409.1556}, 2014.

\bibitem{deep-fscale-multi-task-17}
Vishwanath~A. Sindagi and Vishal~M. Patel.
\newblock Cnn-based cascaded multi-task learning of high-level prior and
  density estimation for crowd counting.
\newblock In {\em Proceedings of International Conference on Advanced Video and
  Signal Based Surveillance}, pages 1--6, 2017.

\bibitem{deep-multi-pyramid-cnn-17}
Vishwanath~A. Sindagi and Vishal~M. Patel.
\newblock Generating high-quality crowd density maps using contextual pyramid
  cnns.
\newblock In {\em Proceedings of International Conference on Computer Vision},
  pages 1879--1888, 2017.

\bibitem{deep-fscal-pooling-PaDNet-18}
Yukun Tian, Yimei Lei, Junping Zhang, and James~Z. Wang.
\newblock Padnet: Pan-density crowd counting.
\newblock {\em CoRR}, abs/1811.02805, 2018.

\bibitem{d-adaboost-full-05}
Paul Viola, Michael~J Jones, and Daniel Snow.
\newblock Detecting pedestrians using patterns of motion and appearance.
\newblock {\em International Journal of Computer Vision}, 63(2):153--161, 2005.

\bibitem{deep-multi-cnn-boosting-16}
Elad Walach and Lior Wolf.
\newblock Learning to count with {CNN} boosting.
\newblock In {\em Proceedings of European Conference on Computer Vision}, pages
  660--676, 2016.

\bibitem{d-part-11}
Meng Wang and Xiaogang Wang.
\newblock Automatic adaptation of a generic pedestrian detector to a specific
  traffic scene.
\newblock In {\em Proceedings of IEEE Conference on Computer Vision and Pattern
  Recognition}, pages 3401--3408, 2011.

\bibitem{ssim}
Zhou Wang, Alan~C Bovik, Hamid~R Sheikh, and Eero~P Simoncelli.
\newblock Image quality assessment: from error visibility to structural
  similarity.
\newblock {\em IEEE Transactions on Image Processing}, 13(4):600--612, 2004.

\bibitem{deep-pooling-fscale-Defense-18}
Ze Wang, Zehao Xiao, Kai Xie, Qiang Qiu, Xiantong Zhen, and Xianbin Cao.
\newblock In defense of single-column networks for crowd counting.
\newblock In {\em Proceedings of British Machine Vision Conference}, page~78,
  2018.

\bibitem{deep-multi-fusion-Adaptive-18}
Xingjiao Wu, Yingbin Zheng, Hao Ye, Wenxin Hu, Jing Yang, and Liang He.
\newblock Adaptive scenario discovery for crowd counting.
\newblock {\em CoRR}, abs/1812.02393, 2018.

\bibitem{deep-multi-block-mscnn-17}
Lingke Zeng, Xiangmin Xu, Bolun Cai, Suo Qiu, and Tong Zhang.
\newblock Multi-scale convolutional neural networks for crowd counting.
\newblock In {\em Proceedings of International Conference on Image Processing},
  pages 465--469, 2017.

\bibitem{deep-crowd-scene-15}
Cong Zhang, Hongsheng Li, Xiaogang Wang, and Xiaokang Yang.
\newblock Cross-scene crowd counting via deep convolutional neural networks.
\newblock In {\em Proceedings of IEEE Conference on Computer Vision and Pattern
  Recognition}, pages 833--841, 2015.

\bibitem{deep-fscale-multitask-SaCNN-18}
Lu Zhang, Miaojing Shi, and Qiaobo Chen.
\newblock Crowd counting via scale-adaptive convolutional neural network.
\newblock In {\em Proceedings of Winter Conference on Applications of Computer
  Vision}, pages 1113--1121, 2018.

\bibitem{agad-18}
Xiaolin Zhang, Yunchao Wei, Jiashi Feng, Yi Yang, and Thomas~S Huang.
\newblock Adversarial complementary learning for weakly supervised object
  localization.
\newblock In {\em Proceedings of the IEEE Conference on Computer Vision and
  Pattern Recognition}, pages 1325--1334, 2018.

\bibitem{deep-geo-Att-head-18}
Youmei Zhang, Chunluan Zhou, Faliang Chang, and Alex~C. Kot.
\newblock Attention to head locations for crowd counting.
\newblock {\em CoRR}, abs/1806.10287, 2018.

\bibitem{MCNN-16}
Yingying Zhang, Desen Zhou, Siqin Chen, Shenghua Gao, and Yi Ma.
\newblock Single-image crowd counting via multi-column convolutional neural
  network.
\newblock In {\em Proceedings of IEEE Conference on Computer Vision and Pattern
  Recognition}, pages 589--597, 2016.

\bibitem{d-full-bayesian-08}
Tao Zhao, Ram Nevatia, and Bo Wu.
\newblock Segmentation and tracking of multiple humans in crowded environments.
\newblock {\em IEEE Transactions on Pattern Analysis and Machine Intelligence},
  30(7):1198--1211, 2008.

\bibitem{cam-16}
Bolei Zhou, Aditya Khosla, Agata Lapedriza, Aude Oliva, and Antonio Torralba.
\newblock Learning deep features for discriminative localization.
\newblock In {\em Proceedings of the IEEE Conference on Computer Vision and
  Pattern Recognition}, pages 2921--2929, 2016.

\end{thebibliography}
}

\end{document}